# Deep Learning Methods in Neuroscience: From Modeling Molecular Mechanisms to Classifying States of Consciousness

Elena Benderskaya[1[0000-0002-7733-416X]], Anastasiia Alifanova[1[0009-0004-3019-1626]], Svetlana Batalova[1], Vasilisa Zhuk[1] and Anna Kovalenko[1]

[1] Peter the Great St. Petersburg Polytechnic University, Polytechnicheskaya, 29, 195251 St.Petersburg, Russia

helen.bend@gmail.com, anastasiiaalifanova@gmail.com

## Abstract

A critical analysis of contemporary approaches to the study of conscious states using deep learning algorithms, computational neuromodeling and neurobiology was conducted. The review focuses on methods of classification, clustering and modeling of brain states under anesthesia. Moreover, the identification of measurable neurobiological characteristics of brain function is considered significant. This analysis is based on the most significant publications from 2018–2025 from the journals Nature, Communications Biology and Scientific Reports, which were selected according to predefined criteria. The analyzed studies cover neuromodeling with the use of deep learning, as well as the estimation of dynamic functional connectivity and the development of multilevel models of consciousness. A comparative analysis was carried out in the following three major areas: automatic detection of states of consciousness using neural networks based on EEG and fMRI data; modeling of the structural-functional dynamics of the brain under the effects of anesthetics; and detection of neurophysiological indicators which correlate with the level of consciousness.

The obtained conclusions demonstrate the growing effectiveness of deep neural models in the classification and prediction of brain states and the analysis of dynamic structural-functional connectivity. Nonetheless, significant limitations were also identified, including the limited interpretability of the models, the lack of standardized metrics, and the problem of the specificity of consciousness markers. Our findings support the need for developing hybrid, physiologically grounded architectures, which are able to generalize across different biological species (rodents, primates) and clinical groups. Furthermore, such approaches may improve the translational potential of computational models in clinical neuroscience.

Diverse methods of machine learning and computational modeling have become indispensable tools in the neuroscience of consciousness. These methods have demonstrated their effectiveness in tasks of automatic clustering and classification of brain states (global accuracy reaches up to 91%), the development of multilevel models and the identification of connectivity patterns correlated with levels of consciousness. A larger-scale analysis and a larger dataset, as well as the implementation of model interpretability approaches are required for the practical application of the analyzed models. The models based on EEG and LFP are the most promising for clinical application due to their availability and the possibility of real-time monitoring.

## Materials and Methods

For the analysis, all scientific publications meeting the following criteria were selected:

1) the article was published in a scientific journal;
2) the article was written in 2018 or later;
3) the article in English was published in journals of the Nature family;
4) the work is devoted to research on determining the presence of consciousness;
5) the article cites previous studies from related fields;
6) the article contains research that the authors conducted independently based on real data;
7) the work has practical applications;
8) the full text of the article is available.

---

## Introduction

Understanding the neurobiological basis of consciousness and developing objective methods for assessing its level remain among the most challenging tasks in modern brain science. This problem has fundamental theoretical significance and direct clinical implications, especially in the context of diagnosing disorders of consciousness, monitoring the depth of anesthesia, and predicting outcomes in patients with brain injuries. Defining consciousness is a key issue in both philosophy and neurobiology. Many authors define consciousness as a subjective phenomenon that "is lost during anesthesia and restored upon awakening"[6]. Others define consciousness as "the heuristic activity of the human brain"[11]. There are definitions of consciousness based on behavioral scales (for example, the Glasgow Coma Scale). Over the years, numerous methods for modeling and defining consciousness have been proposed; among Russian researchers one of the most prominent is E.A. Umryukhin, who developed the concept of a cybernetic model that described complex global brain functions based on a quantitative analysis of conscious and unconscious mental processes during problem-solving[12], and in his dissertation described his experience in constructing a model of the brain's foresight mechanism.

Today, the rapid development of deep learning methods is opening up new prospects for analyzing complex patterns of brain activity obtained using electroencephalography (EEG), functional magnetic resonance imaging (fMRI), and local field potential (LFP) recordings. This review analyzes some of the most significant studies reflecting current research trends in which these technologies are used to classify states and their transitions, to model the effects of anesthetics, and to search for fundamental principles of the organization of consciousness, such as the temporal dynamics of transient activity in large-scale brain networks[1, 2, 7]. An analysis of these works reveals not only achievements but also systemic methodological problems, ranging from the opacity of neural networks' "black boxes" to the limitations of markers, manifested in their insensitivity to the presence of subjective experience (e.g., dreams) at a stable level of sedation[1, 8]. A critical analysis of these achievements and contradictions is an urgent research priority.

The aim of this work is to conduct a comprehensive critical analysis of modern deep learning and computational modeling methods for clustering states of consciousness, as presented in the most relevant contemporary studies, in order to assess their potential and identify fundamental limitations. To this end, it is necessary to systematize the main methodological approaches by major research areas and conduct a comparative analysis of their applicability, interpretability, and biological plausibility, paying particular attention to the results obtained and the contradictions identified.

**Structure of the Article**

The review begins with an abstract and an introduction; the main body then analyzes three areas of research: tools for automatic state classification, multilevel computational frameworks, and models of dynamic structural-functional connectivity as a marker of consciousness. The conclusion summarizes the results and outlines prospects for practical application.

## 1. Classification of Sources: Three Methodological Approaches

The studies under analysis can be logically grouped into three interrelated but methodologically distinct areas.

**Deep Learning Algorithms for Automatic Classification of Brain States and Detection of Transitions.**

This study proposes a solution for determining brain states and transitions between them [1]. It consists of an architecture comprising two convolutional neural networks and a self-training autoencoder with multimodal clustering (Figure 1).

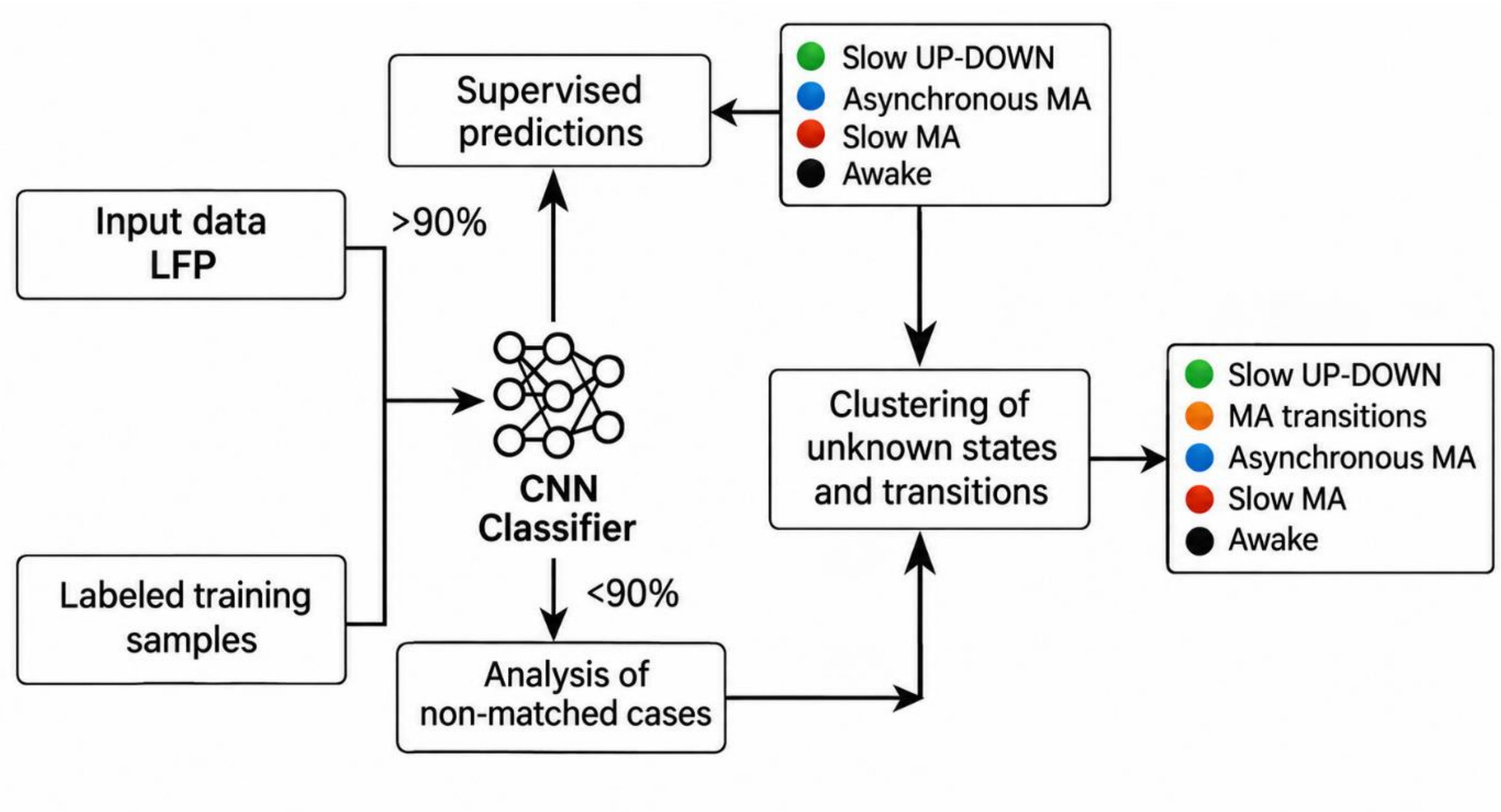

Fig 1. Overview of the state classification using an exclusion strategy [1].

The first CNN classifies the main states (awake (AW), slow-wave sleep (SO), and microarousals (MA)), while the second classifies subtypes of microarousals (slow and asynchronous microarousals (MA)). A key advantage is the introduction of a confidence threshold (CL = 90%), which ensures high model accuracy: samples classified with a confidence level above the threshold are automatically labeled, while ambiguous cases, marked as “unknown,” are sent for further processing by an autoencoder, which reconstructs the signal and analyzes the power spectral density (PSD) for final clustering and identification of transitional states (Figure 2). The method demonstrates high global accuracy (91%) and an average accuracy of 75% ± 6% for transition state detection on local field potential (LFP) data from rats (n = 4), using a “leave-one-out” scheme.

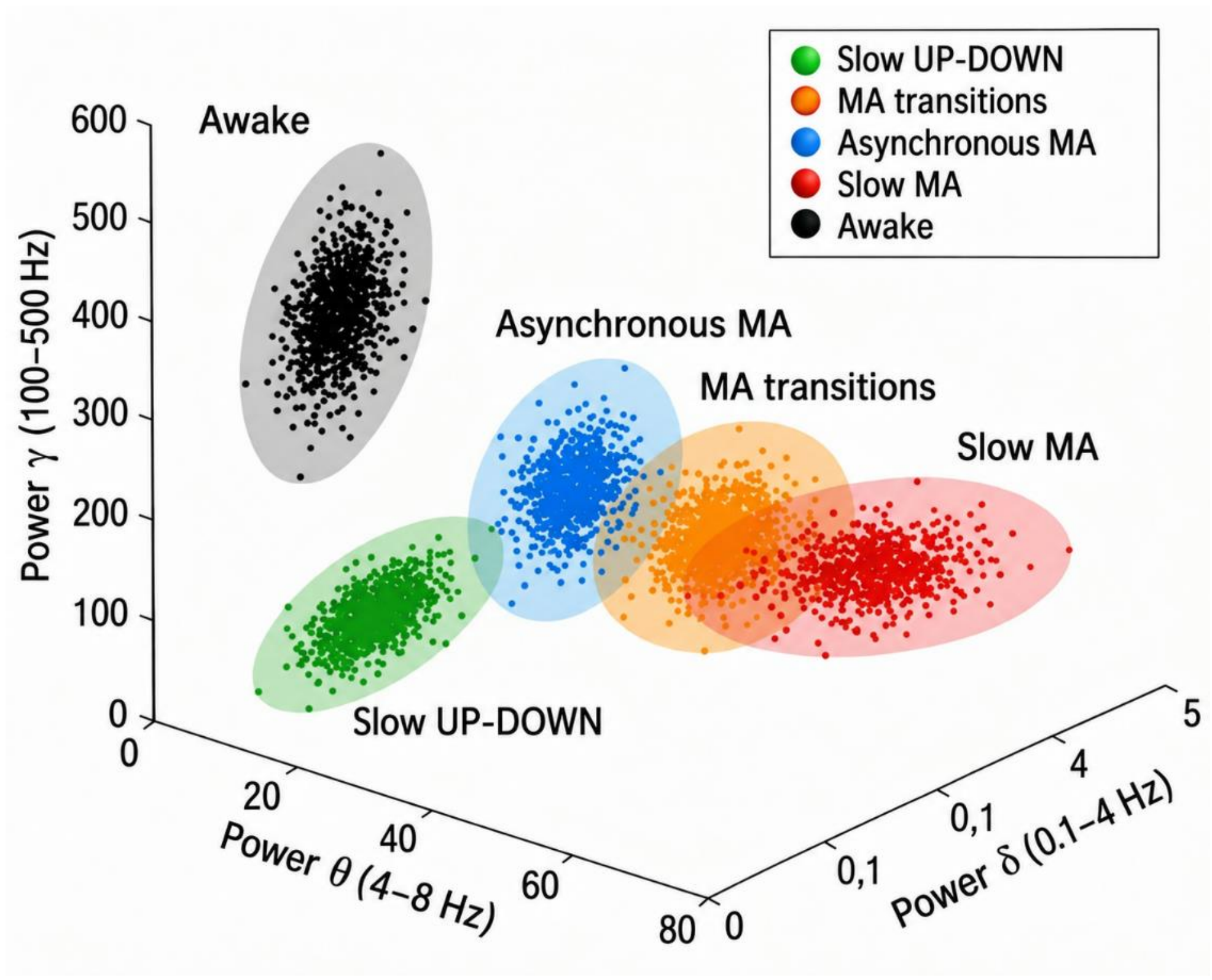

Fig 2. Clustered state samples with transitions [1].

This tool is well suited for automating analysis and is particularly useful for studying dynamic transitions during emergence from anesthesia. However, the approach has significant limitations: it remains a “black box” with low interpretability of neurophysiological features, is trained on data that is limited both in volume (n=4) and type (a single type of LFP), and requires expert tuning of hyperparameters for application in various conditions.

**Multilevel computational models for simulating effects on the brain.**

Unlike approaches focused on classifying pre-existing data, this field aims to create biologically interpretable computational models capable of predicting brain dynamics in response to interventions[2, 6].

Sacha et al.[2] implemented a multilevel model that scales from the molecular level to whole-brain activity. Its architecture is constructed starting from the microlevel, represented by a spiking neuron model (AdEx), through the mesolevel in the form of a spiking neural network, to the entire system - a biophysically grounded mean-field model. This model condenses the dynamics of neural populations into a compact, low-dimensional description, which, in turn, allows for the construction of a model of the entire brain as a network of interacting nodes (mean-field models for each brain region), structurally connected via an individual human anatomical connectome and implemented on The Virtual Brain (TVB) platform.

This approach allows for the direct incorporation of the molecular effects of anesthetics (e.g., propofol and ketamine) by modifying synaptic parameters (the time constants $\tau i$ and $\tau e$ of GABA-A and NMDA receptors, respectively) and the prediction of the emergence, on a macroscopic scale, of the slow-wave activity (UP/DOWN states) characteristic of deep anesthesia. The mean-field formulation makes it possible to accomplish the previously difficult task of creating a large-scale model of the entire brain (at the macroscopic level) with single-cell resolution (at the microscopic level). Limitations include simplified modeling of drug effects (through parameter changes rather than full molecular interactions) and dependence on the quality of the input data.

Another approach involves the use of statistical models to analyze the brain's overall activity. Studies use fMRI (functional magnetic resonance imaging) data from volunteers and employ various statistical methods to analyze the data. For example, using intraclass correlation (ICC), it has been demonstrated in practice that under the influence of anesthesia, the brain loses its "uniqueness"; principal component analysis (PCA) made it possible to reduce the dimensionality of the data and compare the patterns of functional connections between humans and macaques, and dominance analysis helped determine which specific factors had the greatest impact on the results[6]. Statistical methods address the issue of data transparency and interpretability. These methods allowed the authors[6] to identify the most important factors and their relative influence. This approach is well-suited for studies with small, homogeneous samples and serves as a foundation for subsequently building a computational model based on the connectome and generalizing to other species (mice, marmosets)[10].

Studies [2, 6] aim to understand states of consciousness by using anesthesia as a specific switch between states. However, these studies are based on different approaches: in one case, a "bottom-up" approach, where the investigation proceeded from a single molecule to the entire brain structure; in the other, a "top-down" approach, where general patterns were studied at the macrolevel.

**Dynamic functional connectivity and neurovascular dynamics as markers of consciousness.**

The state of consciousness is reflected in the dynamics of interaction between different brain regions[3, 4, 5, 7].

Variational autoencoders (VAE) have become the foundation for analyzing patterns of brain activity in macaques—both while conscious and under anesthesia—to identify hidden patterns. Using this approach, it is much easier to identify previously hidden patterns on fMRI, analyze dynamic functional connections, and investigate how brain activity changes over time and how transitions from one “state” of consciousness to another occur. Because the variational autoencoder extracts important information, this makes it easier to interpret data obtained from clinical trials, which can simplify research.

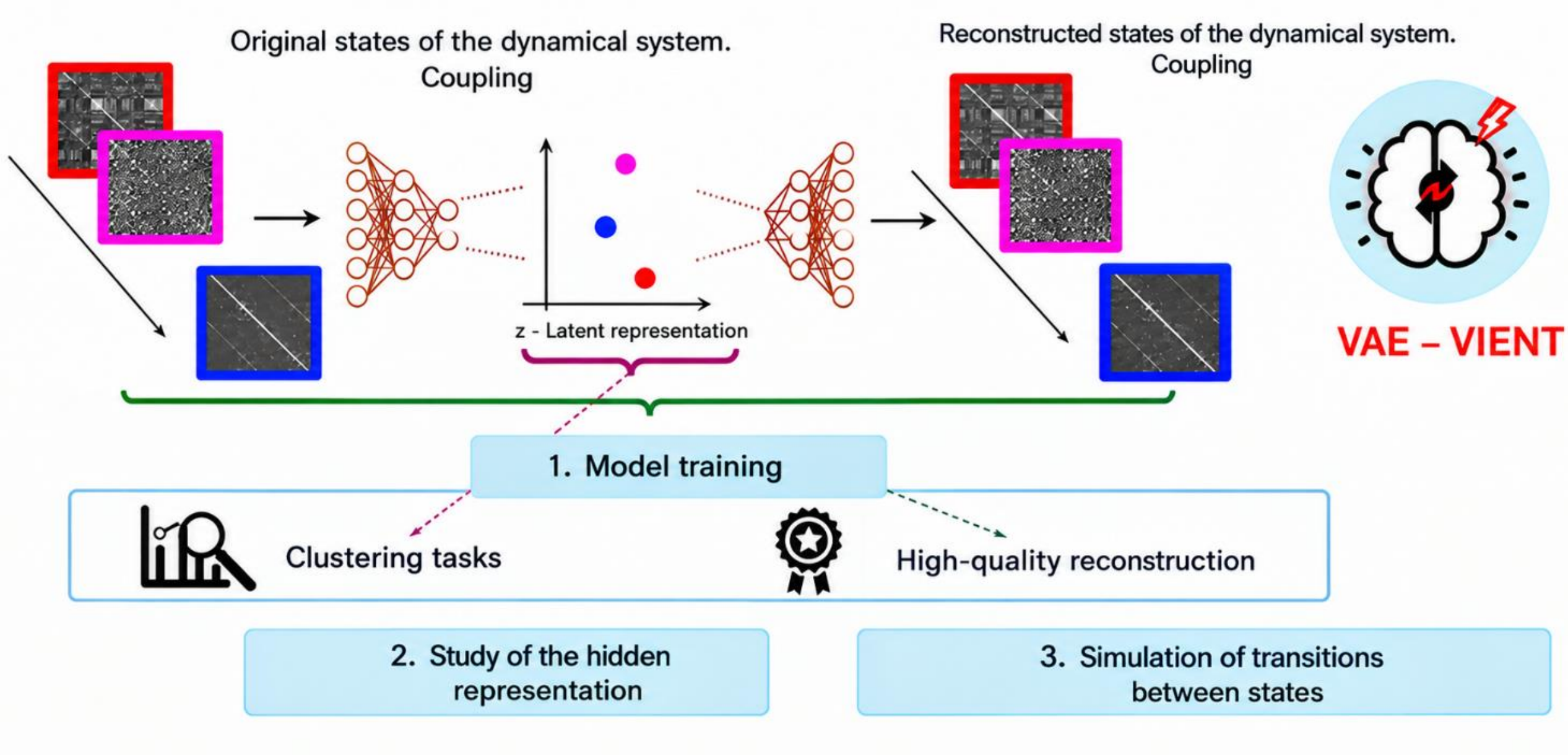

Fig. 3 Illustration of the proposed VAE-VIENT framework. A VAE learns 2D latent representations z = ($z_{1i}$, $z_{2i}$) from dynamic functional connectivity matrices (dFCs). [3]

The latent representation allows for the exploration of both discrete and continuous representations. It also enables the implementation of two simulation paradigms: receptive field analysis, which generates tensor representations to study the perturbation effect of input dFCs, and ablation studies of Global Neuronal Workspace (GNW) connections to investigate the transition from wakefulness to an unconscious state.

Patients with disorders of consciousness show impaired perception of oneself and the external environment is impaired. By applying clustering of EEG connectivity patterns to 237 subjects with consciousness disorders and 101 healthy volunteers, five main patterns of functional brain connectivity were identified, each of which correlated with a person’s level of consciousness[4]. Patterns with high weighted entropy values were characteristic of most healthy individuals, while those with low values were characteristic of patients with severe consciousness disorders. Significant differences in weighted entropy were observed between the control group and all other patient groups (Table 1).

Table 1 - The difference between the weighted entropy values of the control group and the rest of the patient groups [4]

| Patient Group | Difference in WE values compared to healthy patients |
|---|---|
| Patients with minimally conscious state (MCS) | -0.01141 ± 0.00254 |
| Patients with unresponsive wakefulness syndrome (UWS) | -0.01521 ± 0.00250 |
| Patients in an acute condition (Acute) | -0.02627 ± 0.00440 |

The clustering of EEG connectivity patterns was performed using the k-means algorithm; this was employed as the primary method and identified five key “brain states” that served as biomarkers of consciousness level. This yielded an objective, quantitative result. The probability of finding different groups of patients in each state is presented below (Figure 4).

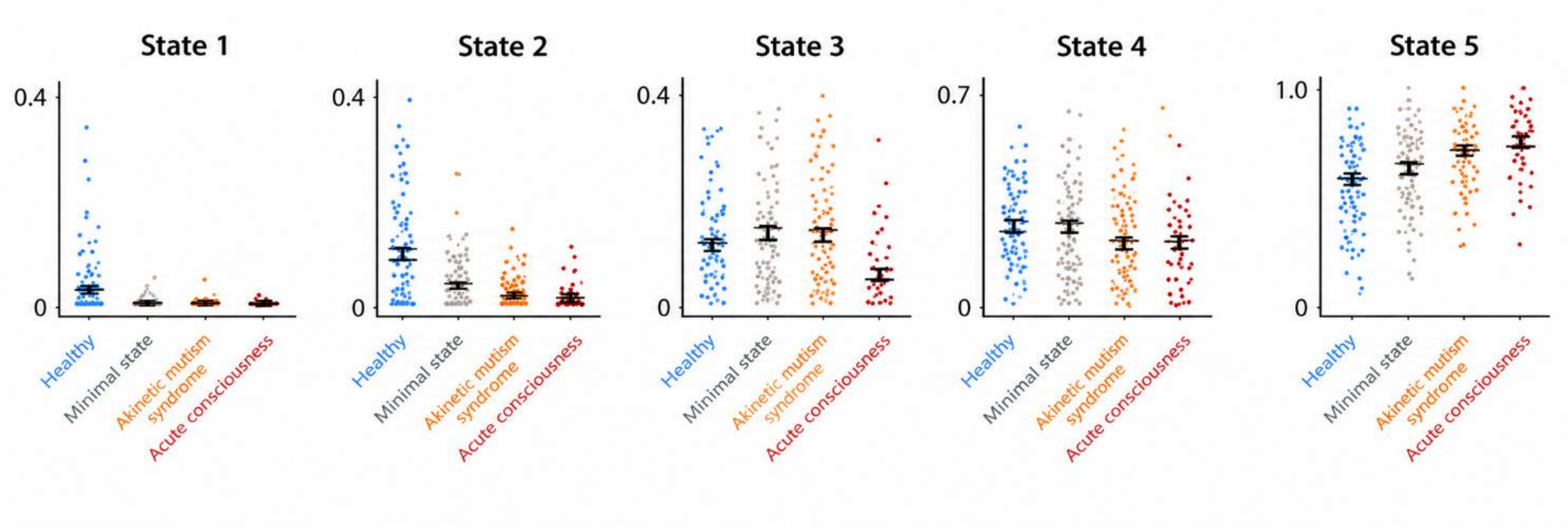

Fig. 4 The probability of finding different groups of patients in each condition [4]

When diagnosing patients’ states, weighted entropy was calculated: first, the results of continuous brain electrical activity were recorded for 20 minutes, and then the probabilities of each recording belonging to a particular cluster for a given individual were calculated. Weighted entropy was then determined based on these probabilities. Based on specific values, it was possible to determine which group a patient belonged to. Additionally, by calculating the weighted entropy coefficient for a single patient over time, it is possible to determine whether the patient is improving or their condition is deteriorating.

A study on neurovascular dynamics reveals that loss of consciousness is accompanied not only by a decrease in the strength of functional connections but also by a global reduction in cerebral blood flow, particularly in the brainstem, thalamus, and cerebellum[5]. This points to a deep connection between neural activity, metabolism, and hemodynamics during changes in consciousness.

To assess the applicability of findings from animal studies, a study of macaque brains during sedation with various anesthetics was considered[7]. In this study, the authors conclude that the primary indicators of unconsciousness are brain regions common to both macaques and humans. Although the experiments in this study were conducted on only five monkeys, these experiments also point to an important finding: during deep general anesthesia with propofol and sevoflurane, brain regions such as the anterior default mode network (aDMN), the secondary visual network, the thalamus, and the anterior cerebellum are simultaneously activated (Figure 5). Although monkeys, unlike humans, lack certain brain networks—such as the attention, language, anterior salience, and visuospatial networks (Figure 6)—the authors emphasize that the study conducted on macaques is consistent with the results obtained from analyses of the human brain.

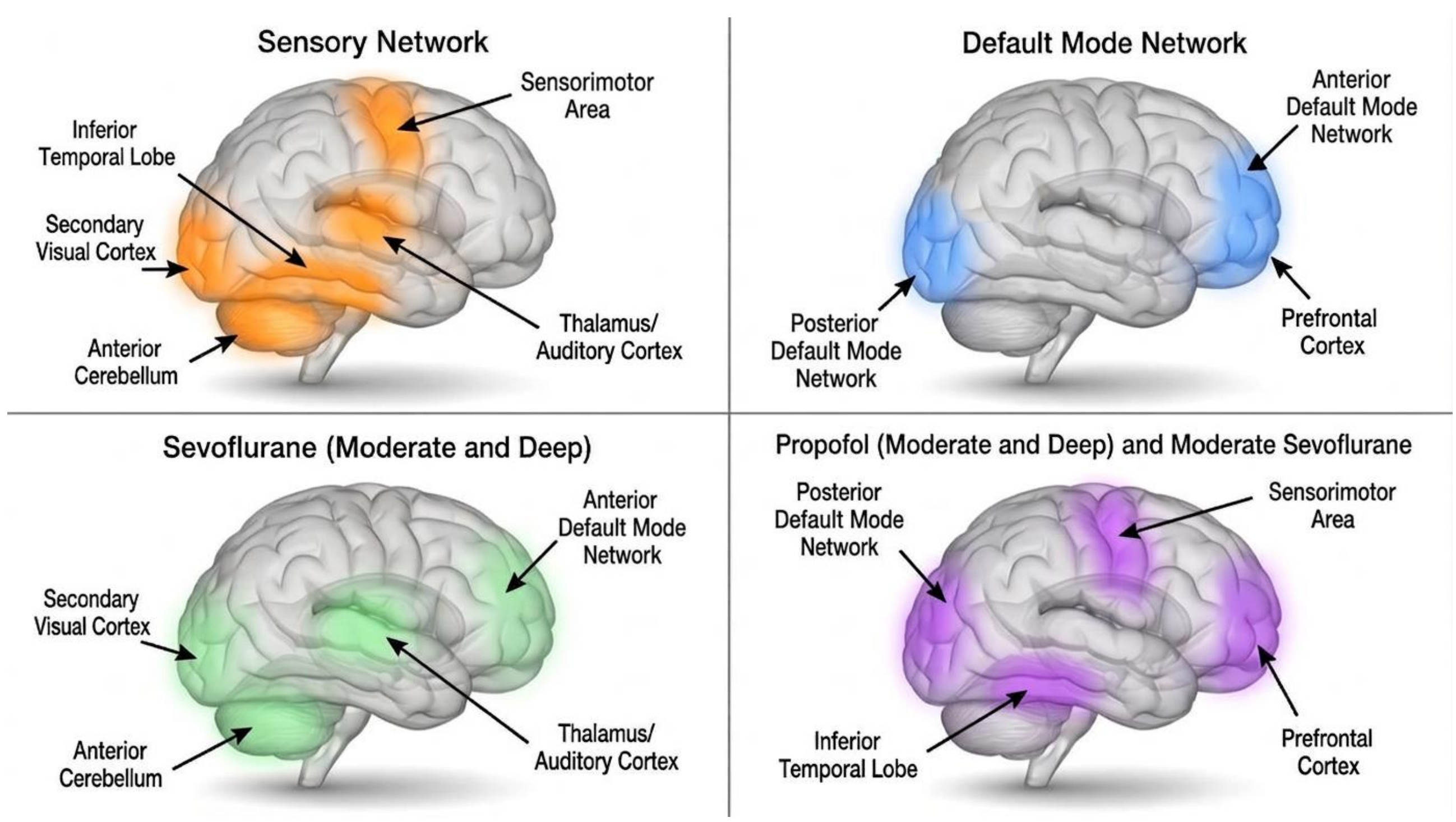

Fig. 5 iCAP clustering according to their temporal overlap [7].

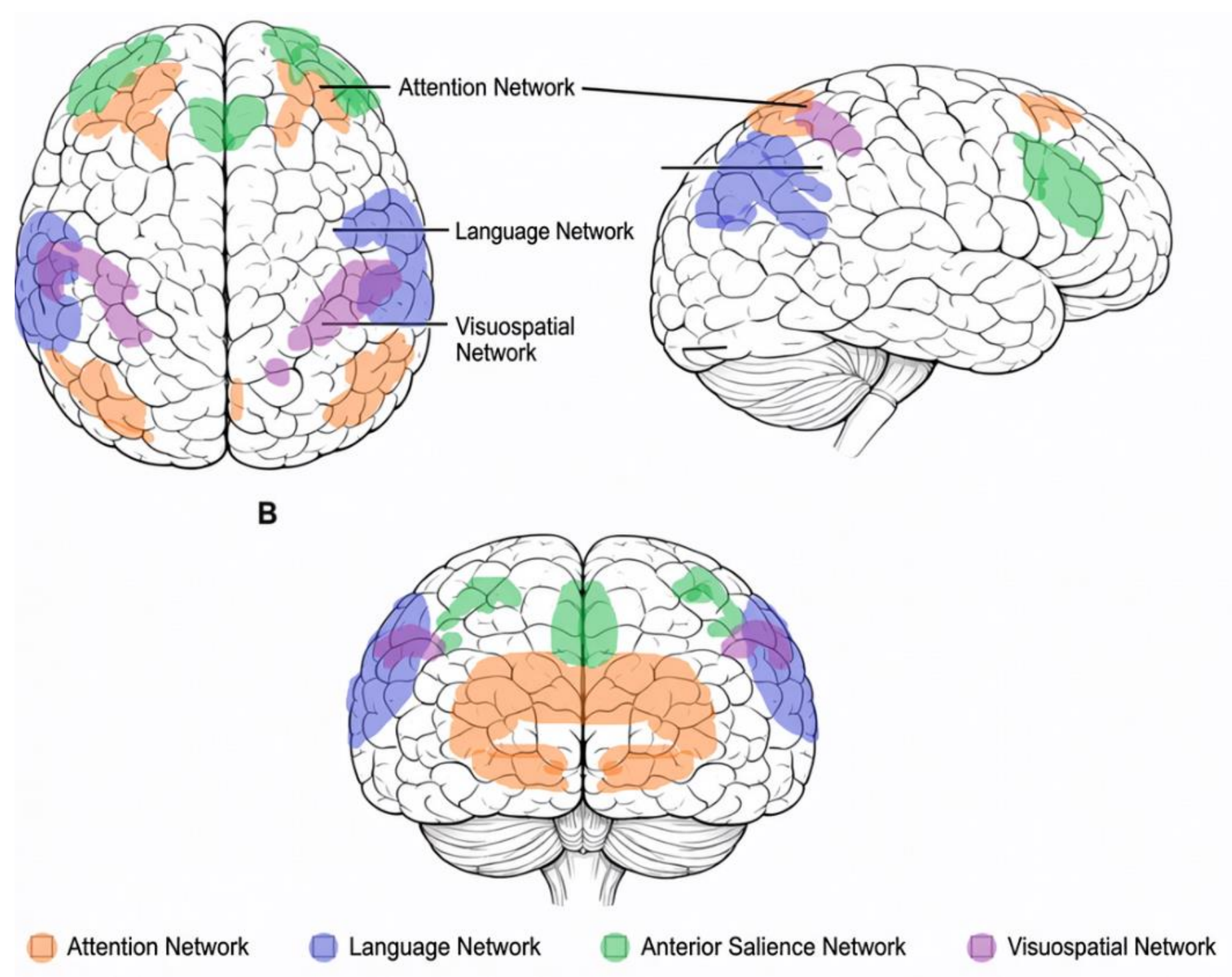

Fig. 6 Brain networks identified exclusively in humans — Attention network (orange), Language network (blue), Anterior salience network (green), Visuospatial network (purple). Reproduced from the original figure published by the authors of the study [7].

## 2. A Critical Analysis of Approaches and the Identification of Gaps

Despite the new findings, the studies presented here have a number of common limitations and issues.

The deep learning methods discussed demonstrate high classification accuracy but often fail to provide a clear neurobiological explanation of exactly which features (frequency, spatial, temporal) underlie the decision[1, 3]. Furthermore, due to the difficulties in collecting data for the datasets, the sample size is often quite small, and it is unknown which features the neural network considered important - a critical barrier in the context of potential clinical applications (e.g., anesthesia monitoring).

In this regard, the computational models that have been developed possess a more transparent internal logic[2, 6]. However, their predictive power and biological validity still need to be tested on large, diverse datasets.

Most models are trained and validated on relatively small and specific datasets: a single animal species (rats, macaques), a single type of anesthetic, data from a single research center, and a small sample size[1]. For example, studies on the “erasure” of unique patterns under the influence of anesthesia have been conducted only for two specific anesthetics and one specific group of volunteers (healthy men); consequently, the generalizability to other groups and drugs requires verification. There are insufficient studies directly comparing the effectiveness of the same algorithms on different types of brain activity recordings (EEG, fMRI, LFP).

A study on dreaming during anesthesia calls into question the simplistic view of the relationship between EEG signal complexity and the presence of subjective experience (dreams)[8]. Using repeated awakenings of participants, the researchers demonstrated that, although measures of complexity such as the state transition complexity index (PCIst) and single-channel Lempel–Ziv complexity (LZc) always decrease during the transition from wakefulness to sedation, they do not differ between episodes of sedation in which patients reported dreaming and episodes without such reports (Figure 7). This significant finding suggests that the aforementioned EEG complexity measures may be sensitive to the subject’s overall level of arousal, but not to the presence or content of conscious experience as such. Furthermore, a significant portion of the participants’ reports regarding the presence or absence of dream experiences could not be accurately interpreted, as in 54% of the responses, participants had difficulty articulating their experiences upon waking.

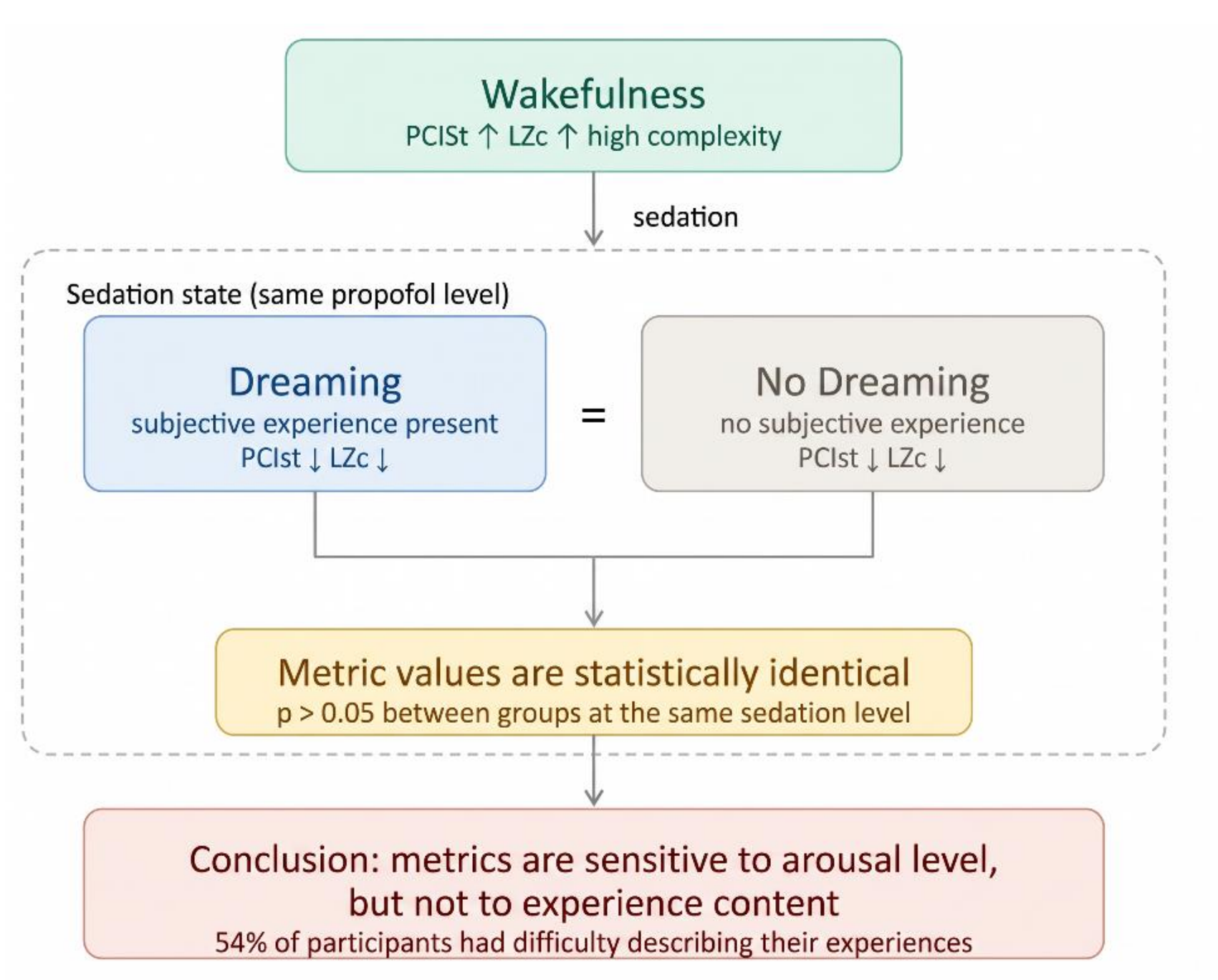

Fig. 7 Conceptual diagram of the experimental results [8].

## 3. Identifying Contradictions and Raising New Questions

An analysis of these models reveals not only gaps but also potential contradictions that point the way for future research.

Researchers point to common features of the unconscious state: a reduction in the complexity and diversity of connectivity patterns and an increased dependence of functional activity on anatomical structure[3, 4, 5]. Studies show that changes in the spatial organization of functional connections are not random[4, 6, 7]. The patterns that contribute most to identifiability during wakefulness are more susceptible to changes under the influence of anesthesia. However, all of this is subject to methodological limitations: different analytical methods and approaches to data interpretation can lead to differing results. For example, variability in approaches to analyzing functional connections can influence conclusions about identifiability. The sample also affects the identifiability of markers: due to ethical approval requirements, the sample is often small and consists of a homogeneous group of subjects, which significantly averages out the results and fails to account for, among other things, individual differences in how anesthesia is tolerated.

Findings from studies on dreaming during sedation demonstrate that even in the same patient, the same level of anesthesia (particularly with propofol sedation) can result in different experiences regarding dreaming[8]. This calls into question the possibility of identifying a universal indicator of consciousness, since scientists have observed vivid dream episodes in subjects even in states traditionally attributed to unconsciousness.

Neurovascular dynamics must also be taken into account when studying functional connectivity[5]. Changes in blood flow do not merely passively follow neural activity but may have their own dynamics that influence measurable fMRI signals and, possibly, the functional capabilities of neural ensembles. This raises questions about approaches that focus exclusively on synaptic and cellular dynamics, as they model neural activity but do not account for metabolic constraints and the influence of blood flow on functional connectivity[2].

## Conclusions

Taken together, the studies reviewed here demonstrate that machine learning and computational modeling have become indispensable tools in the neuroscience of consciousness. They have proven effective in tasks such as the automatic classification of brain states, the construction of multilevel models illustrating the effects of anesthetics, and the identification of connectivity patterns that correlate with the level of consciousness[1, 2, 3, 4, 5, 6, 7]. A key general conclusion is that the conscious state is associated with high diversity, complexity, and flexibility of the brain's structural-functional connections, whereas unconscious states are associated with reduced sensitivity to external stimuli, and functional connectivity becomes more closely tied to the underlying structural architecture.

To create a reliable simulation of a brain model, it is necessary to take into account not only functional connectivity and neuronal activity but also neurovascular dynamics[2].

The focus should be on identifying specific patterns of neuronal activity rather than a single indicator of consciousness[8]. Future research should shift from a simple "consciousness/unconsciousness" dichotomy to an analysis of the dynamics of neural processes over time, as well as the search for markers of consciousness that correlate with the presence of dream experiences and take other factors into account.

To apply the studied models more effectively in practice, larger-scale research and a broader sample size are needed, as well as the implementation of model interpretation. Methods based on EEG and LFP are the closest to clinical implementation, thanks to their accessibility and the ability to monitor in real time[1, 4]. They could form the basis for decision-support systems in the diagnosis of consciousness disorders or for more accurate monitoring of sedation depth in the intensive care unit and the operating room. However, before this can happen, large-scale studies must be conducted to demonstrate that the proposed algorithms not only correlate with the current state but also have prognostic value for patient outcomes. Computational models also have the potential for screening and predicting transitions between states of consciousness[2, 6].